\pdfoutput=1

\documentclass[11pt]{article}

\usepackage[]{coling}

\usepackage{times}
\usepackage{latexsym}

\usepackage[T1]{fontenc}

\usepackage[utf8]{inputenc}

\usepackage{microtype}

\usepackage{inconsolata}

\usepackage{graphicx}

\usepackage{cite}
\usepackage{amsmath,amssymb,amsfonts}
\usepackage{algorithmic}
\usepackage{textcomp}
\usepackage{xcolor}
\usepackage{bbm}

\usepackage{booktabs}
\usepackage{soul}
\usepackage{subcaption}
\usepackage{multicol}
\usepackage{tabularx}
\usepackage{enumitem}

\title{Towards Robust and Accurate Stability Estimation of Local Surrogate Models in Text-based Explainable AI}

\author{Christopher Burger \\
\footnotesize
  University of Mississippi\\
  \footnotesize
  \texttt{cburger@olemiss.edu} \\\And
    Charles Walter\\
    \footnotesize
  University of Mississippi\\
\footnotesize
  \texttt{cwwalter@olemiss.edu}\\\And
  Thai Le \\
  \footnotesize
  Indiana University  \\
  \footnotesize
  \texttt{tle@iu.edu} \\\And
  Lingwei Chen \\
  \footnotesize
  Wright State University \\
  \footnotesize
  \texttt{lingwei.chen@wright.edu}}

\begin{document}
\maketitle
\begin{abstract}
Recent work has investigated the concept of adversarial attacks on explainable AI (XAI) in the NLP domain with a focus on examining the vulnerability of local surrogate methods such as \textsc{Lime} to adversarial perturbations or small changes on the input of a machine learning (ML) model. In such attacks, the generated explanation is manipulated while the meaning and structure of the original input remain similar under the ML model. Such attacks are especially alarming when XAI is used as a basis for decision making (e.g., prescribing drugs based on AI medical predictors) or for legal action (e.g., legal dispute involving AI software). Although weaknesses across many XAI methods have been shown to exist, the reasons behind why remain little explored. Central to this XAI manipulation is the similarity measure used to calculate how one explanation differs from another. A poor choice of similarity measure can lead to erroneous conclusions about the stability or adversarial robustness of an XAI method. Therefore, this work investigates a variety of similarity measures designed for text-based ranked lists referenced in related work to determine their comparative suitability for use. We find that many measures are overly sensitive, resulting in erroneous estimates of stability. We then propose a weighting scheme for text-based data that incorporates the synonymity between the features within an explanation, providing more accurate estimates of the actual weakness of XAI methods to adversarial examples. \textit{Source codes will be made publicly available.}

\end{abstract}

\section{Introduction}

Continuous advancements in AI/ML have enabled the proliferation of powerful, yet complex, models in many aspects of our daily lives \citep{dong2021survey}. Although these models are deployed across a wide range of fields, disciplines like medicine or finance, where the consequences of failure are especially high, must be particularly cautious about model errors.
When such a failure occurs, understanding its nature and type becomes critical. For example, a model built for bird classification that misidentifies a pine warbler as a yellow-throated vireo may cause little more than embarrassment. However, a collision detection system that interprets a pedestrian in a crosswalk as background noise could lead to catastrophic failure. In any case, it is essential for its stakeholders to be informed of why such failure happens. With ever-increasing model complexity, their inner workings often become impenetrable black boxes. To address this
challenge, the discipline of explanatory AI (XAI) \citep{tjoa2020survey,dwivedi2023explainable} has emerged to develop tools that allow both developers and users of models to understand \textit{why a model works the way it does.}

A common method in XAI is to generate a simplified, inherently interpretable model, often a constrained regression or decision tree, as an approximation of the complex model \citep{guidotti2018survey}. These approximations, known as surrogate models, are then used to gain insight into why the original model has made certain predictions. Among surrogate models, global surrogates attempt to explain the model as a whole, but if they were sufficiently accurate, they could be used directly over the complex model. In contrast, local surrogates \citep{garreau2020explaining} are widely preferred due to their increased fidelity, as they focus on explaining an individual prediction (or at most a small subset). Nevertheless, surrogate models \textit{are} algorithms themselves, and may be subject to the same issues as the models they are designed to explain \citep{InterpretationNN}. 

\paragraph{XAI Stability.} One such issue is stability or robustness, where an insignificant change in input should result in a correspondingly insignificant change in output \citep{alvarezmelis2018robustness}. This property is leveraged in adversarial attacks, where a model that is unstable can return obviously erroneous results, given small changes to the input. Local surrogate models have been shown to lack stability in many types of data, including image, tabular, and text~\citep{slack2020fooling,ivankay2022fooling,xaifooler}. That is, the explanation of the complex model can be appreciably different given an insignificant change in the input. If an XAI model cannot produce consistent explanations across very similar documents when the output of the complex model itself remains similar, then the explanations from the XAI method become suspect. Suspect explanations \textit{cannot be trusted}, so the original complex model remains opaque and can be prevented from use due to legal or social ramifications despite superior efficacy over other models. This will seriously slow the advance of AI and its adoption in practice.

\paragraph{Similarity Measures in XAI.} Previous work on XAI stability, such as \citep{InterpretationNN,slack2020fooling,ivankay2022fooling} has focused on the \textit{existence} of instability given some fixed maximum amount of change applied to the input. This contrasts to standard adversarial attacks on classification models. These attacks often also have some perturbation limit, but the process has a clearly defined end point, which is a change in the predicted class. Previous work in XAI stability generally uses a similar approach to that of standard adversarial attacks, but lacks this clearly defined metric of success, and instead terminates at search exhaustion or perturbation limit \citep{xaifooler}.

As the existence of instability has already been demonstrated, we seek to refine this knowledge by asking \textit{``what is the importance of the similarity measure used to guide the adversarial search process?''}.
The similarity measure is the engine for selection of appropriate perturbations in the search process and directly controls the acceptance of perturbations and the overall determination of success or failure of the algorithm.  
In particular we ask: Are certain similarity measures superior to others in providing quality adversarial explanations or accurate results in terms of XAI robustness? If so, under what conditions? Specifically, how do different measures compare in terms of \textit{sensitivity}, or the propensity to show a difference in similarity with respect to another measure. Table \ref{tab:comparative_perturbations_example} provides an illustration. It shows a perturbed explanation with 75\% similarity to the original calculated using measure RBO$_{0.5}$ (to be defined in Section \ref{measure_def}) is compared with eight other measures, resulting in up to 65\% lower similarity between documents solely from choosing a new measure.

\begin{table}[tb]
    \footnotesize

    
    

    
    \begin{subtable}[t]{.48\textwidth}
    \raggedright
    \setlength\tabcolsep{7 pt}
        \begin{tabular}{clc|clc}
           \multicolumn{3}{c}{\textbf{Original Explanation}} & \multicolumn{3}{c}{\textbf{Perturbed Explanation}} \vspace{2pt}
            \\
             \multicolumn{2}{c}{\textbf{Feature}} & \multicolumn{1}{c}{\textbf{Weight}}&  \multicolumn{2}{c}{\textbf{Feature}} & \multicolumn{1}{c}{\textbf{Weight}}  \\
            \hline \\
            1 & heartburn & 1.77 & 1 & heartburn &  2.16 \\
                2 & eat & 0.59 & 2 & choking & 0.35 \\
                3 & vomit & 0.35 & 3 & puked & 0.34 \\
                4 & choking & 0.26 & 4 & eat & 0.33 \\
                5 & feel & 0.17 & 5 & like & 0.11 \\
                6 & lot & 0.15 & 6 & pain & 0.09 \\
                & ... & & &  ... & \\
                
            \hline \\
        \end{tabular}
\end{subtable}%

\begin{subtable}[t]{.48\textwidth}
        \caption*{\textbf{Comparative Similarities}}
        \raggedright
            \begin{tabular}{@{\extracolsep{4pt}}cccc}
                \toprule
                
                    \multicolumn{1}{c}{\textbf{RBO}$_{0.7}$}
                    & \multicolumn{1}{c}{\textbf{RBO}$_{0.9}$}
                    & \multicolumn{1}{c}{\textbf{Jaccard}}
                    & \multicolumn{1}{c}{\textbf{Jaccard}$_w$}
                    \vspace{2pt}
                    \\
                    0.69 & 0.74 & 0.72 & 0.90 
                    \vspace{4pt}
                    \\
                    \multicolumn{1}{c}{\textbf{Kendall}}
                    & \multicolumn{1}{c}{\textbf{Kendall}$_w$}
                    & \multicolumn{1}{c}{\textbf{Spearman}}
                    & \multicolumn{1}{c}{\textbf{Spearman}$_w$}
                    \vspace{2pt}
                    \\
                     0.10 & 0.48 & 0.50 & 0.50 \\
               \bottomrule
            \end{tabular}
    \end{subtable}%
    \caption{Explanation changes under similarity measure \textbf{RBO$_{0.5}$} with final similarity of \textbf{0.75} (out of maximum 1.0) only from a few text perturbations: ``{I have a \textcolor{blue}{\textbf{\textit{lot}}} of heartburn and I feel like I'm choking when I eat. I also have a lot of stomach pain and I \textcolor{blue}{\textbf{\textit{vomit}}} a lot.}''$\rightarrow$ ``{I have a \textcolor{red}{\textbf{\textit{lots}}} of heartburn and I feel like I'm choking when I eat. I also have a lot of stomach pain and I \textcolor{red}{\textbf{\textit{puked}}} a lot.}''}
    \label{tab:comparative_perturbations_example}
\end{table}

Given that such substantial differences can exist, and subsequently alter the conclusions about the suitability of an XAI method, we use this as motivation to build purpose-driven variants for similarity of text-based ranked lists. We elect to alter the similarity measure directly using synonymity weighting, where a perturbed word and its original are compared using an inner measure that incorporates their synonymity. Standard measures operate on \textit{strict equality between discrete features} (words in our case) in the lists being compared. This discards valuable information contained within the structure of the language itself. The removal of this information is not congruent with our focus of maintaining the close meaning of the perturbed word, as the similarity measure should reflect the end goal of the adversarial process by accounting for the ``closeness'' among the words. 

\paragraph{Our contributions are summarized as follows.} 
\begin{enumerate}[leftmargin=\dimexpr\parindent-0.2\labelwidth\relax,noitemsep]
    \item Exploring the effect on the quality of stability estimates using different similarity measures to guide the adversarial search process in text-based XAI. Moreover, identifying measures that are unsuitable for use due to excessive sensitivity resulting in exaggerated indications of instability.
    \item The extension of common measures to use synonymity weighting focused on providing a more appropriate choice for adversarial methods in XAI. We leverage prior work in element-weighted similarity to allow a more accurate comparison between explanations, which provides a superior understanding of the robustness of an XAI method.
    \item A comprehensive comparison against adversarial examples previously determined to be successful at some threshold of similarity under the standard measures. We show that prior conclusions about XAI instability drawn using certain existing measures may be inaccurate, especially for methods with naive search procedures.

\end{enumerate}


\section{Background and Related Work}

Previous work on XAI stability has focused on evaluating models using tabular or image data in various interpretation methods, which often use small perturbations of the input data to generate appreciably different explanations \citep{alvarezmelis2018robustness,InterpretationNN,alvarezmelis2018robustness}, or generate explanations consisting of arbitrary features \citep{slack2020fooling}. Garreau et al. showed that key features can be omitted from the resulting explanations by changing parameters and that artifacts of the explanation generation process could produce misleading explanations \citep{garreau2020explaining,garreau2022looking}. This was further extended to the analysis to text data \citep{mardaoui2021analysis} but only with respect to fidelity instead of stability of surrogate models.

This work then focuses on the least explored domain, text. Previous work directly involving adversarial perturbations for XAI exists but has focused on determining the existence of such perturbations rather than establishing which components in the XAI method are most vulnerable \citep{sinha2021perturbing,ivankay2022fooling,xaifooler}. 
Other relevant work in the text domain includes ~\citet{ivankay2022fooling}, which utilized a gradient-based approach but assumed white-box access to the target model; and \citet{sinha2021perturbing}, which generated adversarial attacks against black-box XAI methods. However, their experiment design may have led to an underestimation of stability as explored in \citet{xaifooler} which investigated the inherent instability of the method of choice's (\textsc{Lime}) sampling process for text data, and provided an alternate search strategy focused on the preferential perturbation of features deemed unimportant.

\subsection{XAI Method}
While there are many XAI methods available, we narrow our choices by focusing on two criteria: a level of generalizability to other XAI methods and the explanations generated are to satisfy the prior desiderata that constitute an effective explanation. Namely, concision, order, and weight \citep{xaifooler}. From these criteria we choose Local Interpretable Model-agnostic Explanations (\textsc{Lime})~\citet{LIME_Ribeiro} as our target explanatory algorithm.
\textsc{Lime} is a commonly used and referenced tool in XAI frameworks, which has been integrated into critical ML applications such as finance~\citep{gramegna2021shap} and healthcare~\citep{kumarakulasinghe2020evaluating,fuhrman2022review}. To explain a prediction, \textsc{Lime} trains a shallow, inherently explainable surrogate model such as Logistic Regression on training examples that are synthesized within a neighborhood of an individual prediction. The resulting explanation is an ordered collection of features and their weights from this surrogate model that satisfies our requirements for a quality explanation outlined above. For text data, explanations generated by \textsc{Lime} have features that are individual words contained within the document to be explained, which can be easily understood even by non-specialists. We note that our method is applicable to any XAI method that returns an explanation in this format, satisfying our criteria of generalizability.

We emphasize that our goal is to investigate the importance of the similarity measure used to guide the adversarial process. \textsc{Lime} is not the focus of our inquiry, but is used as a familiar standard due to its wide adoption of previous work in adversarial XAI.

\setlength{\abovedisplayskip}{2pt}
\setlength{\belowdisplayskip}{2pt}

\section{Testing the Effects of Alternate Similarity Measures}

\subsection{Adversarial XAI Algorithm}

\paragraph{Adversarial Goal.} As our goal is fundamentally similar to prior work in that we seek the discovery of perturbations that induce instability in XAI methods, our constraints when generating adversarial perturbations are generally equivalent to prior work that has established the existence of instability. That is, the \textit{ meaning} of the input is retained, as well as an identical predicted class for the perturbed input in the original model. We use the general perturbation process described in \citet{xaifooler} with one addition. While we retain a maximum amount of perturbations, we also use a threshold of similarity, $\tau$, used to signify a successful attack. 

\paragraph{Searching for Perturbations.} The general search process is focused on the comparison of explanations and not on locating appropriate perturbations. We use the greedy search procedure standard to previous work, where the indices of the words within the original document are ordered by importance (or lack thereof) to the model to be explained. The importance is calculated by removing the word at the $i^{th}$ index and determining the change in the predicted probability. These indices are sorted, filtered by the constraints, and then iterated through where the word at index $i$ has perturbations generated to replace it. These perturbations are the $n$ nearest neighbors in some embedding space where the final replacement is chosen by the largest decrease in similarity. 

\subsection{Similarity Measures} \label{measure_def} 
We assume the explanations generated are ranked lists, ordered by importance to the surrogate model (as is standard for \textsc{Lime} and common with other XAI methods). Then it is natural to choose measures for similarity or distance designed specifically for ranked lists. From the candidates, we select four popular measures that represent two major focuses of comparison, set-based overlap, and dissonance between paired features. For the following definitions, let $\mathbf{A}, \mathbf{B}$ denote the ranked lists. 

\noindent \textbf{Jaccard Index} is the ratio of the size of the nonempty intersection of the lists (here viewed as sets) over the size of their union (Eq. \ref{Jaccard}). If the intersection is empty, the resulting similarity is defined to be zero. 

\begin{equation}\label{Jaccard}
    J(\mathbf{A},\mathbf{B}) = \frac{|\mathbf{A} \cap \mathbf{B}|}{|\mathbf{A} \cup \mathbf{B}|}
\end{equation}

\noindent \textbf{Kendall's Tau Rank Distance} (Eq. \ref{eq:Kendall}) counts the number of pairwise inversions between $\mathbf{A}$ and $\mathbf{B}$ (where $\mathbbm{1}[\cdot]$ is the indicator function. To allow for comparison of lists unequal in size, we assume that all excess elements of the larger list are automatically disjoint.
\begin{equation}\label{eq:Kendall}
    \sum_{i=1}^{\textbf{min}(|\mathbf{A}|,|\mathbf{B}|)} (\mathbbm{1}[\mathbf{A}[i] \neq \mathbf{B}[i]]) +  | |\mathbf{A}| - |\mathbf{B}| |
\end{equation}

\noindent \textbf{Spearman's footrule} (Eq. \ref{eq:Spearman}) is the sum of the difference between the location $i$ of each feature $a \in \mathbf{A}$ to its corresponding location $j$ in $\mathbf{B}$. Spearman's footrule is effectively the $L_1$ distance between ranked lists. Similarly to Kendall's Tau, the footrule is by default not suitable for disjoint lists, but the footrule distance is bounded, with a maximum total distance of $\lfloor \frac{|\mathbf{A}|^2}{2}\rfloor$ with an individual element having at most $|\mathbf{A}| - 1$ possible distance. This leads to a natural choice for a penalty value for missing elements of $\frac{|\mathbf{A}|}{2}$, since for two completely disjoint lists we have $\sum_{i = 1}^{|\mathbf{A}|}\frac{|\mathbf{A}|}{2} = \frac{|\mathbf{A}|^2}{2}$. Using a penalty, the total maximum footrule distance can increase (Appendix \ref{apn:spearman}), but we note that this adjustment is not required as the proportion of disjoint elements between explanations is often small. As such, the original bound remains useful even without the added penalty factor, which may avoid concerns of an ill-justified penalty. 

\begin{equation}\label{eq:Spearman}
  \sum_{a \in \mathbf{A}} | i - j |
\end{equation}
 

\noindent \textbf{Rank-biased Overlap (RBO)} is a summation of successively larger intersections, each weighted by a term in a convergent series. This weighting scheme is controlled by a parameter $p \in (0,1)$ that can be adjusted to ascribe more or less weight to the top-k features. In general, values further down the list are weighted as less significant, which is often the case in XAI, as only the top few features are of interest to many end users \citep{verma2020counterfactual}.
RBO \citep{10.1145/1852102.1852106} is defined in Eq. \ref{eq:RBO} where 
 $d$ is the current depth of the ranking, and $k$ is the maximum depth. 
\begin{equation}\label{eq:RBO}
  RBO(\mathbf{A},\mathbf{B},p) =  (1-p)  \sum_{d = 1}^{k}{p^{d-1}}  \frac{|\mathbf{A}_{:d} \cap \mathbf{B}_{:d}|}{d}
\end{equation}

 \newcolumntype{Y}{>{\centering\arraybackslash}X}
\begin{table*}[h]
\footnotesize
\begin{tabularx}{\textwidth}{c *{11}{Y}}
\toprule
\multicolumn{1}{c}{\textbf{}}
& \multicolumn{1}{c}{\textbf{$\tau$}}
& \multicolumn{1}{c}{\textbf{RBO}$_{0.5}$}
& \multicolumn{1}{c}{\textbf{RBO}$_{0.7}$}
& \multicolumn{1}{c}{\textbf{RBO}$_{0.9}$}
& \multicolumn{1}{c}{\textbf{Jaccard}}
& \multicolumn{1}{c}{\textbf{Jaccard}$_w$}
& \multicolumn{1}{c}{\textbf{Kendall}}
& \multicolumn{1}{c}{\textbf{Kendall}$_w$}
& \multicolumn{1}{c}{\textbf{Spearman}}
& \multicolumn{1}{c}{\textbf{Spearman}$_w$}\\
\cmidrule(lr){2-11}
& 30\%  & 0.12 &  0 &  0 & 0.02 & 0 & 0.95  & 0.52  & 0.14  & 0.02\\
\rotatebox[origin=c]{90}{\textbf{GB}}& 40\%  & 0.26 &  0.07 &  0 & 0.24 & 0 & 0.98  & 0.64  & 0.38  & 0.12\\
& 50\%  & 0.40 &  0.29 & 0.07 & 0.88 & 0 & 1  & 0.74  & 0.83  & 0.43\\
& 60\%  & 0.40 & 0.40 &  0.28 & 1 & 0.05 & 1  & 0.81  & 1  & 0.69\\
  \midrule

& 30\%  & 0.06 &  0.02 &  0 & 0.06 & 0 & 1  & 0.42  & 0.18  & 0.04\\
\rotatebox[origin=c]{90}{\textbf{S2D}}& 40\%  & 0.18 &  0.04 &  0 & 0.52 & 0 & 1  & 0.48  & 0.58  & 0.28\\
& 50\%  & 0.24 &  0.2 &  0.08 & 0.98 & 0.02 & 1  & 0.72  & 0.92  & 0.60\\
& 60\%  & 0.24 &  0.3 &  0.28 & 1 & 0.14 & 1  & 0.84  & 1  & 0.94
\\
\bottomrule 
\end{tabularx}
\caption{Attack success rates on \textsc{Lime} under DistilBERT.}
\label{tab:success_rates}
\end{table*}

\subsection{Experimental Setup}\label{sub:experimental_setup}
To provide the raw material for our analysis, we generate batches of 50 adversarial examples using the algorithm in \citet{xaifooler}. Each batch is generated with respect to a similarity measure and a success threshold. Nine similarity measures are used: The Jaccard index, Kendall's Tau Rank Distance, and Spearman's footrule, each in their standard and weighted implementations and RBO, with weighting parameters 0.5, 0.7, and 0.9. Combined with success thresholds ($\tau$) of 30\%, 40\%, 50\%, and 60\% this results in 1,800 adversarial examples per data set. These thresholds were chosen to provide a balance between a sufficient difference between explanations and retained document quality (see Appendix \ref{apn:example} for a comparative example at different thresholds). For the data sets, we use two of those included in \citet{xaifooler} and their associated pre-trained models. The first is the Twitter gender bias dataset (average 11 words per document)  (\textbf{GB}) \citep{dinan-etal-2020-multi} and the second is the symptoms-to-diagnosis dataset (average 29 words per document)  (\textbf{S2D}) (Kaggle). The final and longest-length dataset, IMDB movie reviews, proved computationally infeasible due to the excessive time requirements to generate the examples, approximately 125 days of continuous computation on a single A6000. The model to be explained is a DistilBERT \citep{sanh2019distilbert} fine-tuned on each respective dataset, which was chosen because it is the model with the least computational overhead. Without loss of generalizability, we choose a single efficient model to explain due to (1) our own computational limitation and (2) our focus is on the general properties of similarity measures and \textit{not their specific performance associated with a given XAI method or model}.

\subsection{Results and Discussion}

Immediately seen is the inappropriateness of certain similarity measures for use in text-based adversarial XAI attacks. Kendall's tau is extremely sensitive with almost 100\% attack success across every combination of threshold and data set (Table \ref{tab:success_rates}). This level of sensitivity renders the measure useless here as the inherent instability of \textsc{Lime} is the cause, and little perturbation-induced instability is present. This follows for the weighted version as well, though the success rates are not quite as high. Jaccard, Spearman, and Spearman$_w$ also show excessive sensitivity for the higher similarity thresholds of 50\% and 60\%. 

Jaccard$_w$ and RBO$_{0.9}$ exhibit the opposite behavior, instead being coarse with few, if any, successful attacks under all but the most lenient threshold. This is not inherently negative, as both measures effectively require the top few characteristics to change, completely removing the explanation in the case of Jaccard$_w$ and at least a substantial decrease in the ranking for RBO$_{0.9}$. Since much of the weight of an explanation is often associated with the top features, these measures may prove useful as a fast heuristic for a substantial explanation difference.

The attack success rates are generally consistent between the datasets. As the sampling rate for \textsc{Lime} was optimized according to previous work, the inherent instability of the sampling process was kept similar despite the size difference.

Ultimately, Kendall and Jaccard measures by default are of limited use due to sensitivity or poor perturbed document quality. RBO in general provides a good balance between sensitivity and perturbed document quality across all thresholds at the cost of requiring manual adjustment of the weighting parameter for the given threshold. The Spearman measures show promise in comparison with RBO for more demanding thresholds but suffer from sensitivity for larger values. Spearman$_w$ may prove particularly useful if the original explanatory weights are deemed important or the fine-tuning of RBO too cumbersome.

\section{Constructing Synonymity Weighted Similarity Measures}
The previous results show that certain measures are ill-suited for use in adversarial XAI in their standard form due to their extreme sensitivity. We now modify these measures to incorporate synonymity between paired features to reduce the sensitivity and so increase the reliability of the conclusions about the robustness of an XAI model to adversarial perturbation.

\subsection{Mappings Between Explanations}\label{sec:mappings}
As our goal is to include the synonymity estimate between paired features $a \in \mathbf{A},b \in \mathbf{B}$ where $\mathbf{A},\mathbf{B}$  are the original and perturbed explanations, respectively, we require a way to determine the pairing $a \to b$ to allow comparison. To do so, we define a mapping using the perturbation process to link elements from both lists.

Creating this mapping is simple if $|\mathbf{A}| = |\mathbf{B}|$ and the perturbation process is restricted to single-word substitutions. Any element $a \in \mathbf{A}$ located at the index $i$ that is perturbed to some value $b$ must be located at some index $j$ in $\mathbf{B}$ or no longer part of the surrogate model and is thus missing from explanation $\mathbf{B}$. For the latter case, this generally means eliminated in importance. 
However, if the search process does not exclude highly ranked features, these can be selected (and often are without careful choice of search constraints), resulting in substantial differences under certain similarity measures despite little substantial difference in meaning. This is one of the major issues that we seek to address with synonymity weighting. 
\subsection{Mappings and Disjoint Explanations}
We note that most perturbed features from standard text XAI perturbation methods remain present in the perturbed explanation, but when constraining the size of the explanation either through the surrogate model itself, or in the case of truncating the resulting explanation to the top k features (done usually to reduce explanation complexity for the end user), we often encounter unpaired elements between the explanations. For measures that rely on consonant lists, an adjustment must be made to allow for dissonant elements. In this case to handle any unpaired elements we apply a penalty value $p$, its value dependent on both the similarity measure and user choice. 

Some measures can be simply extended to handle disjoint elements. For example, Kendall's Tau (Eq. \ref{eq:Kendall}) can be easily extended to the comparison of different sized lists by declaring whatever excess element(s) that exist in the larger list to be automatically dissonant. But even for measures whose structure discards element pairings (generally set intersection-based), we must establish a mapping to apply the synonymity weighting. We will assume that all elements contained within the original explanation $\mathbf{A}$ are mapped, either to an element in $\mathbf{B}$ or to the null mapping, indicating that the measure-specific penalty should be applied. Additionally, multi-word substitutions for the perturbation method increases the difficulty of maintaining appropriate syntactic consistency and have not seen much use in adversarial explanations in XAI. As such, we will assume that all perturbation methods replace at most one word per iteration. 

Now given our mapping, we must decide what form of synonymity weighting to apply.

\subsection{Constructing Weighted Similarity}\label{sec:weightingMethods}
With some mapping established between $\mathbf{A}$ and $\mathbf{B}$, how should the synonymity weighting be implemented? 

We define a function $Syn(a,b) \longrightarrow [0,1]$ where $a, b$ are features within explanations $\mathbf{A}, \mathbf{B}$ respectively. The function $Syn(\cdot)$ returns a value proportional to the synonymity between the features $a$ and $b$. We constrain the definition of $Syn(a,b)$ minimally, with the only condition required being $Syn(x,x) = 1$, where $1$ is the absolute maximum similarity possible. In particular, the interval itself is subject to alteration. We choose $[0,1]$ to provide for a simple representation for the proportion of similarity between two words and so make the analysis easier to conduct, but this is not required and other choices are possible. Of these alternative choices, the interval $[-1,1]$ may be the most intuitive choice by letting $-1$ indicate that $b$ is an antonym of $a$, and $0$ being that $b$ is completely unrelated to $a$. 
\subsection{A Simple Example}
For example, let $a = good$, $b = bad$, and $c = frog$. Now for the interval $[0,1]$, $Syn(a,b) = Syn(a,c) = 0$ as neither word is synonymous with \textit{good}. For the interval $[-1,1]$ $Syn(a,b) \approx -1$ and $Syn(a,c) = 0$ as \textit{bad} is an antonym of the adjective \textit{good}. Since part of speech checking is a commonly imposed constraint in text-based adversarial XAI, we assume that there is no ambiguity between words that span multiple syntactic categories such as \textit{ good} (\textbf{noun}) and \textit{good} (\textbf{adjective}) should the synonymity measure be capable of handling this distinction. In particular, measures that incorporate embedding vectors do not necessarily make a distinction between identical words with multiple meanings (the embeddings are not \textit{ multisense}). These embeddings will also often not produce values with such an obvious delineation. \textit{Good} and \textit{frog} will likely possess some similarity $> 0$ by construction of the embedding. 
We choose for the remainder of the paper the $[0,1]$ interval to simplify the exposition and experimental verification.

\subsection{Creating Weighted Measures}
\label{sec:experiment}
To demonstrate the effectiveness of synonymity weighting,, we modify the four similarity measures defined previously in Section \ref{measure_def}.

\noindent \textbf{Jaccard Index.} We can extend the Jaccard index to use synonymity weighting by using the mapping $M$ and a paired word synonymity function $Syn$ to return the similarity between each word of $\mathbf{A}$ and its corresponding match in $ \mathbf{B}$. Identical words are treated as without weighting in that they are assigned a value of one, while unequal words gain from the synonymity function $Syn$.
\begin{align}\label{Jaccard2}
    J_W(\mathbf{A},\mathbf{B}) = \frac{\displaystyle\sum_{i = 1}^{|\mathbf{A}|}\textit{Syn}(\mathbf{A}[i],M(\mathbf{A}[i]))}{|\mathbf{A} \cup \mathbf{B}|} 
\end{align}

\noindent \textbf{Kendall's Tau.} We extend Kendall's Tau to use synonymity weighting by adjusting the value of a mapped dissonant pair $a,b$ at an equal location by multiplying by $1 - Syn(a,b)$. For highly synonymous replacements ($Syn(a,b) \to 1$) this assigns a distance close to zero for the mapped pair, and for dissimilar words it approaches the default distance. 

\begin{equation}\label{eq:Kendall2}
    \sum_{i=1}^{\textbf{min}(|\mathbf{A}|,|\mathbf{B}|)} (\mathbbm{1}[\mathbf{A}[i] \neq \mathbf{B}[i]]*(1 - Syn(a,b))  
    + d,
\end{equation}
\noindent where $d = | |\mathbf{A}| - |\mathbf{B}| | $

\noindent \textbf{Spearman's Footrule.} We extend the Spearman's footrule to use synonymity weighting by taking the summation of three mutually exclusive conditions a pair of features within the explanations may have. For features unchanged in both explanations, we calculate the distance as normal. For features $a,b$ with a mapping $a \to b$ we calculate the minimum between the standard distance divided by the synonymity between features $Syn(a,b)$ and the maximum possible distance under the default footrule $|A| - 1$. 

\begin{equation}\label{eq:Spearman2}
  \sum_{a \in \mathbf{A\cap B}} | i - j | +  \sum_{\substack{a \in \mathbf{A} \\ b \in \mathbf{B} \\ a \to b}} s + \sum_{a \in \mathbf{A \cap \bar{B}}} p ,
\end{equation}
\noindent where $s = \mathbf{min}(\frac{| i - j | }{Syn(a,b)}, |\mathbf{A}| - 1)$

\noindent \textbf{Rank-biased Overlap.} For RBO we apply the same idea as in Eq. \ref{Jaccard2}  where the size of the intersection is increased by the similarity of each pair of mapped disjoint elements.

\subsection{Empirical Validation}
We reuse the experimental procedure from Section \ref{sub:experimental_setup} with the exception of excluding the weighted versions of the Jaccad Index, Kendall's Tau, and Spearman's footrule. Combining synonymity weighting with the default weighting requires enough subjective decisions in defining the measure that we omit the construction and testing of these hybrid measures here and reserve this for a more substantial discussion in future work.

\begin{figure*}[!ht]

\minipage{\textwidth}
  \centerline{\includegraphics[width=\linewidth,height=0.20\linewidth]{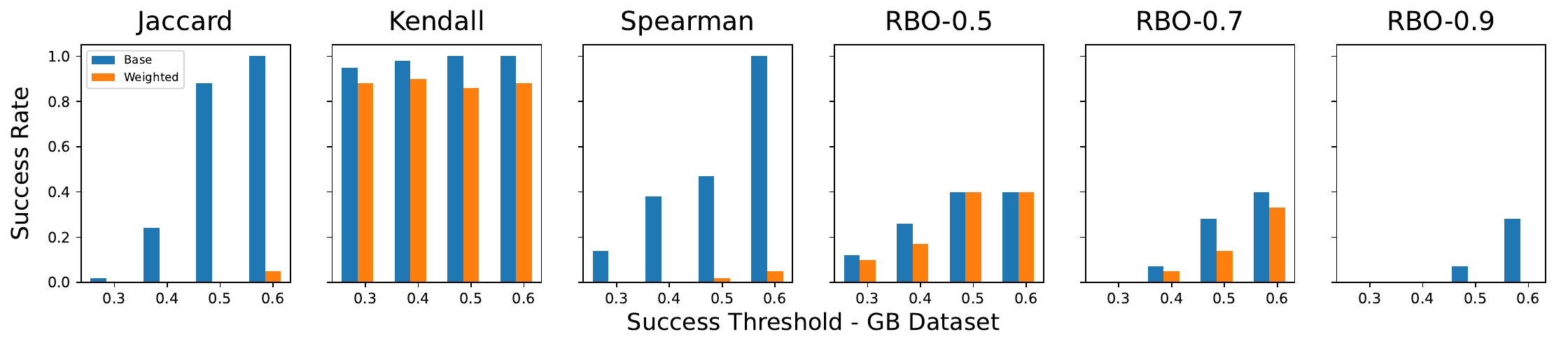}}
  \endminipage
\hfill
\minipage{\textwidth}  \centering
  \centerline{\includegraphics[width=\linewidth,height=0.20\linewidth]{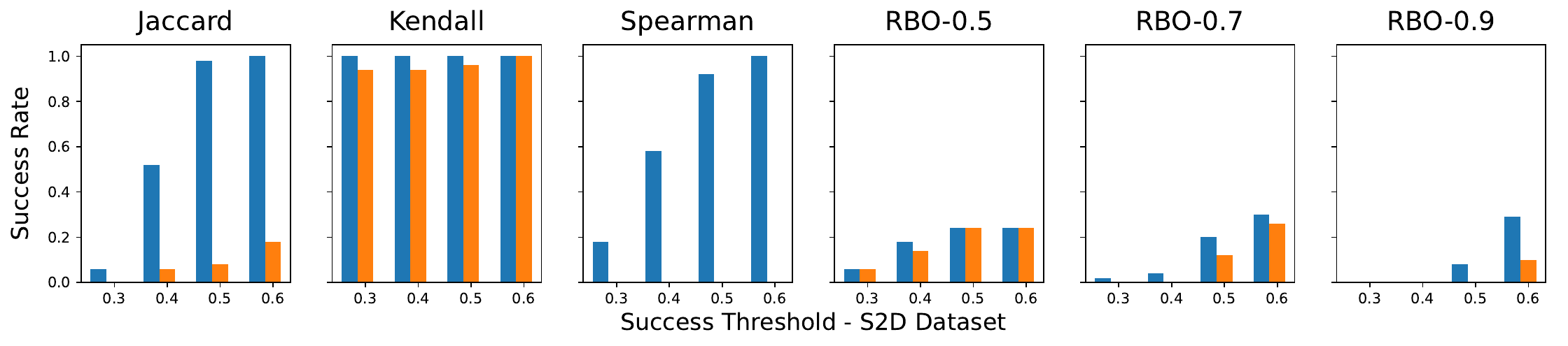}}
  \endminipage
\caption{Successful attack rates under threshold $\tau$ for standard and synonymity weighted explanations (Base Measure \textcolor{blue}{(Blue)} - Synonymity Weighted Measure (\textcolor{orange}{Orange)})}
\label{fig:success}%
\end{figure*}

All the following results are calculated exclusively with respect to successful attacks. We provide data before and after the application of synonymity weighting on the overall attack success rate given some success threshold $\tau$ (Figure \ref{fig:success}).
 The method used to determine feature synonymity is cosine similarity on the GloVe Twitter 27B-25d embedding. 

\begin{table*}[ht]

\footnotesize
\begin{tabularx}{\textwidth}{c *{13}{Y}}
\toprule
\multicolumn{1}{c}{\textbf{}}
& \multicolumn{1}{c}{\textbf{$\tau$}}
& \multicolumn{2}{c}{\textbf{Jaccard}}
& \multicolumn{2}{c}{\textbf{Kendall}}
& \multicolumn{2}{c}{\textbf{Spearman}}
& \multicolumn{2}{c}{\textbf{RBO}$_{0.5}$}
& \multicolumn{2}{c}{\textbf{RBO}$_{0.7}$}
& \multicolumn{2}{c}{\textbf{RBO}$_{0.9}$}\\
\cmidrule(lr){2-14}
& & \textbf{Base} & \textbf{Syn$_w$} & \textbf{Base} & \textbf{Syn$_w$} & \textbf{Base} & \textbf{Syn$_w$} & \textbf{Base} & \textbf{Syn$_w$} & \textbf{Base} & \textbf{Syn$_w$} & \textbf{Base} & \textbf{Syn$_w$}
\\ \addlinespace[3pt]
& 30\%  & 0.02 &  0 & 0.95 & 0.88 & 0.14 & 0 & 0.12  & 0.10   & 0 & 0 &  0 &  0 \\
\rotatebox[origin=c]{90}{\textbf{GB}}& 40\%   &  0.24 & 0  & 0.98 & 0.90 & 0.38  & 0 & 0.26 & 0.17  & 0.07 &  0.05 & 0 & 0 \\
& 50\%  & 0.88 &  0 &  1 & 0.86 & 0.83 & 0.02 & 0.40 & 0.40   & 0.28 & 0.14 &  0.07 &  0  \\
& 60\%  & 1 &  0.05 &  1 & 0.88 & 1 & 0.05 & 0.40 & 0.40    & 0.40 & 0.33 &  0.28 &  0 \\
\midrule
& 30\%  & 0.06 & 0  &  1 & 0.94 & 0.18 & 0 & 0.06 & 0.06   & 0.02 & 0 & 0 &  0 \\
\rotatebox[origin=c]{90}{\textbf{S2D}}& 40\%  & 0.52 &  0.06 &  1 & 0.94 & 0.58 & 0  & 0.18 & 0.14   & 0.04 & 0.04 &  0 &  0 \\
& 50\%  & 0.98 &  0.08 &  1 & 0.96 & 0.92 & 0 & 0.24 & 0.24   & 0.20 & 0.12 &  0.08 &  0 \\
& 60\%  & 1 &  0.18 &  1 & 1 & 1 & 0 & 0.24 & 0.24   & 0.30 & 0.26 & 0.28 & 0.10\\
\bottomrule
\end{tabularx}

\label{tab:success}
\caption{Successful attack rates under threshold $\tau$ for standard and synonymity weighted explanations 
}
\end{table*}

\subsection{Results and Discussion}

\noindent \textbf{Jaccard \& Spearman:} We immediately see the drastic reduction in the attack success rate for the Jaccard Index and Spearman's footrule. For Jaccard under the \textbf{GB} dataset, we see that the three most stringent thresholds (30\%, 40\%, and 50\%) are reduced to zero success, while the 60\% threshold is reduced to 5\%. Similar results hold for the \textbf{S2D} dataset. Spearman is even more sensitive than Jaccard, with every \textbf{S2D} success under the original calculation changed to failure. Spearman's footrule was investigated to determine if the weighting scheme was appropriate due to the enormous success reduction and comparatively more complex weighting formulation over Jaccard. However, the different choices of the penalty value showed little change in the results with only slight ($\leq 5\%$) reductions in the calculated similarity. These results demonstrate that both standard measures are extremely sensitive to changes in the explanation that are due to properties of the measure, and not due to a change in fundamental meaning of the explanation. With such significant decreases in attack success, any conclusions generated about the stability of XAI under these measures should be viewed with suspicion. 

\noindent \textbf{RBO:} In contrast to Jaccard and Spearman, RBO gains much less from synonymity weighting, only showing minor changes in success rate for weighting values of 0.5 and 0.7. For weighting value 0.9 it is not surprising to see more efficacy as a more uniform distribution of importance to each feature. 
The lack of overall success associated with RBO$_{0.9}$ makes a firm judgment on its usefulness premature, data with more successful attacks are required. The appreciable difference in the change in the success rate between RBO and Jaccard was not expected as both are intersection-based. The relatively minor gain of RBO from synonymity weighting is likely due to its own inherent weighting scheme; to confirm this, the intrinsically weighted version of the other measures can be tested, and we leave this for future investigation. Overall, RBO remains a strong candidate for use in adversarial XAI.

\noindent \textbf{Kendall:} Kendall's Tau also appears to gain little from synonymity weighting, with minor changes on \textbf{S2D} and about twice the effectiveness on \textbf{GB}. This conclusion may be premature, as the extreme sensitivity (note the near 100\% success rate across every threshold and dataset) appears to completely overpower the effects of synonymity weighting. 
This measure remains poorly suited for adversarial XAI, and the synonymity weighting tested here is not sufficient to allow much, if any, practical use.
\section{Conclusion}

This work shows that a poor choice in the similarity measure can drastically skew the results of an XAI adversarial attack, either by selecting a measure that is too sensitive and overstating the weakness of the XAI method or by choosing a measure too coarse and overstating the method's resilience. This work also informs that AI practitioners and researchers should choose these measures judiciously, keeping in mind how sensitive a measure is to the inherent variation between explanations. Our proposal of synonymity weighting for such a measure can alleviate some level of sensitivity on standard similarity measures. Synonymity weighting allows for broader classes of measures to be used, whereas without it, measures must be restricted to a subset of inherently weighted measures in order to provide reasonable estimates of XAI robustness. Finally, such a weighting schema also incurs negligible computational overhead as the bottleneck in current algorithms is the explanation generation itself.



\clearpage

\section*{Limitations}
\paragraph{Transferability to other XAI methods:} While the fundamentals of the discussion on similarity measures are not unique to \textsc{Lime}, our general conclusions on the fitness of particular similarity measures may not be applicable to different XAI methods. XAI methods with less inherent instability may see more useful results from measures like Kendall's Tau, where its excessive sensitivity and subsequently exaggerated attack effectiveness may no longer pose a problem. 

\paragraph{Other similarity measures:} The similarity measures used here were designed for the use of ranked lists. Other measures with this purpose or more generalized measures exist, and our conclusions may not be reflective of these untested measures.

\paragraph{Optimality of the Synonymity Weighting Methods:} The construction for the specific synonymity weighting definitions is a necessary first step towards this new research direction and calls for further exploration from the community. Spearman's footrule may benefit from a new method that may prevent the severe change in similarity between the original and weighted versions. Our choice of numerical synonymity is another prime choice for optimization, especially as part of speech checking is not associated with our selection. Superior estimates of synonymity can probably be obtained with alternative methods. 

\section*{Broader Impacts and Ethics Statement}
The authors anticipate that there is no reasonable
cause to believe that the use of this work would cause harm, direct or implicit. The authors disclaim
any conflicts of interest related to this work.

\clearpage
\bibliography{custom,anthology}

\clearpage
\appendix
\onecolumn
\section{Additional Experimental Results}\label{apn:example}
\begin{table*}[!h]
    \caption{Perturbed Explanations at different attack success thresholds. Degradation in textual quality is apparent with continued perturbation. \textit{Italics} within an explanation indicate a word subject to perturbation or its perturbed form.}
    \footnotesize
    \begin{center}
          \textbf{Top-6 Features}
    \end{center}

    \begin{subtable}[h]{.2\textwidth}
        \caption*{Original}
        \raggedright
        \setlength\tabcolsep{3 pt}
            \begin{tabular}{clc|}
                \hline \\
                 & \multicolumn{1}{c}{\textbf{Feature}} & \multicolumn{1}{c}{\textbf{Weight}}  \\
                \hline \\
                1 & headaches & 2.80 \\
                2 & sensitivity & 0.30 \\
                3 & having & 0.26 \\
                4 & \textit{tried} & 0.25 \\
                5 & sound & 0.25 \\
                6 & vomiting & 0.21 \\
                & ... & 
            \end{tabular}
    \end{subtable}%
       \begin{subtable}[h]{.2\textwidth}
        \caption*{60\% Threshold}
        \raggedright
        \setlength\tabcolsep{3 pt}
            \begin{tabular}{clc|}
                \hline \\
                 & \multicolumn{1}{c}{\textbf{Feature}} & \multicolumn{1}{c}{\textbf{Weight}}  \\
                \hline \\
                1 & headaches & 2.85 \\
                2 & counter & 0.30 \\
                3 & sound & 0.25 \\
                4 & don't & 0.23 \\
                5 & sensitivity & 0.19 \\
                6 & having & 0.16 \\
                & ...&
            \end{tabular}
    \end{subtable}%
        \begin{subtable}[h]{.2\textwidth}
        \caption*{50\% Threshold}
        \raggedright
        \setlength\tabcolsep{3 pt}
            \begin{tabular}{clc|}
                \hline\\
                 & \multicolumn{1}{c}{\textbf{Feature}} & \multicolumn{1}{c}{\textbf{Weight}}  \\
                \hline \\
                1 & headaches & 3.01 \\
                2 & having & 0.38 \\
                3 & sensitivity & 0.34 \\
                4 & taking & 0.24 \\
                5 & vomiting & 0.20 \\
                6 & don't &  0.18\\
                & ... & 
            \end{tabular}
    \end{subtable}%
        \begin{subtable}[h]{.2\textwidth}
        \caption*{40\% Threshold}
        \setlength\tabcolsep{3 pt}
        \raggedright
            \begin{tabular}{clc|}
                \hline\\
                 & \multicolumn{1}{c}{\textbf{Feature}} & \multicolumn{1}{c}{\textbf{Weight}}  \\
                \hline \\
                1 & headaches & 2.86 \\
                2 & sensitivity & 0.30 \\
                3 & having & 0.25 \\
                4 & \textit{strived} & 0.20 \\
                5 & counter & 0.17 \\
                6 & light & 0.17 \\
                & ... & \\
            \end{tabular}
    \end{subtable}%
        \begin{subtable}[h]{.2\textwidth}
        \caption*{30\% Threshold}
        \setlength\tabcolsep{3 pt}
        \raggedright
            \begin{tabular}{clc}
                \hline \\
                 & \multicolumn{1}{c}{\textbf{Feature}} & \multicolumn{1}{c}{\textbf{Weight}}  \\
                \hline \\
                1 & headaches & 3.06 \\
                2 & sensitivity & 0.27 \\
                3 & \textit{strived} & 0.26 \\
                4 & left & 0.25 \\
                5 & painful & 0.22 \\
                6 & counter & 0.21 \\
                & ... &
            \end{tabular}
    \end{subtable}%
    \hrule
    \vspace{3pt}
    \begin{center}
         \textbf{New Locations for original Top-6 Features} 
    \end{center}
    \hrule
    
    \begin{subtable}[h]{.2\textwidth}
        \raggedright
        \setlength\tabcolsep{3 pt}
            \begin{tabular}{llc|}
                 & \multicolumn{1}{c}{\textbf{Feature}} & \multicolumn{1}{c}{\textbf{Weight}}  \\
                \hline \\
                1 & headaches & 2.80 \\
                2 & sensitivity & 0.30 \\
                3 & \textit{having} & 0.26 \\
                4 & \textit{tried} & 0.25 \\
                5 & sound & 0.25 \\
                6 & vomiting & 0.21 \\
                & ... & 
            \end{tabular}
    \end{subtable}%
       \begin{subtable}[h]{.2\textwidth}
        \raggedright
        \setlength\tabcolsep{3 pt}
            \begin{tabular}{llc|}
                 & \multicolumn{1}{c}{\textbf{Feature}} & \multicolumn{1}{c}{\textbf{Weight}}  \\
                \hline \\
                1 & headaches & 3.10 \\
                5 & sensitivity & 0.19 \\
                6 & having & 0.16 \\
                9 & tried & 0.14 \\
                3 & sound & 0.25 \\
                8 & vomiting & 0.15 \\
                & ... & 

            \end{tabular}
    \end{subtable}%
        \begin{subtable}[h]{.2\textwidth}
        \raggedright
        \setlength\tabcolsep{3 pt}
            \begin{tabular}{llc|}
                 & \multicolumn{1}{c}{\textbf{Feature}} & \multicolumn{1}{c}{\textbf{Weight}}  \\
                \hline \\
                1 & headaches & 3.01 \\
                3 & sensitivity & 0.34 \\
                2 & having & 0.38 \\
                12 & tried & 0.26 \\
                13 & sound & 0.0.08 \\
                5 & vomiting & 0.20 \\
                & ... & 
            \end{tabular}
    \end{subtable}%
        \begin{subtable}[h]{.2\textwidth}
        \setlength\tabcolsep{3 pt}
        \raggedright
            \begin{tabular}{llc|}
                 & \multicolumn{1}{c}{\textbf{Feature}} & \multicolumn{1}{c}{\textbf{Weight}}  \\
                \hline \\
                1 & headaches & 3.06 \\
                2 & sensitivity & 0.27 \\
                13 & having & 0.25 \\
                3 & \textit{strived} & 0.20 \\
                16 & sound & 0.15 \\
                13 & vomiting & 0.10 \\
                & ... & 
            \end{tabular}
    \end{subtable}%
        \begin{subtable}[h]{.2\textwidth}
        \setlength\tabcolsep{3 pt}
        \raggedright
            \begin{tabular}{llc}
                 & \multicolumn{1}{c}{\textbf{Feature}} & \multicolumn{1}{c}{\textbf{Weight}}  \\
                \hline \\
                1 & headaches & 3.06 \\
                2 & sensitivity & 0.27 \\
                13 & \textit{assuming} & 0.22 \\
                3 & \textit{strived} & 0.26 \\
                10 & sound & 0.15 \\
                18 & vomiting & 0.04 \\
                & ... & 
            \end{tabular}
    \end{subtable}%

    \hrule
\vspace{5pt}
\begin{center}
\textbf{Original Text} 
\end{center}

I have been having headaches for a while now. They are usually on the left side of my head and are very painful. I also get nausea, vomiting, and sensitivity to light and sound. I have tried taking over-the-counter pain relievers, but they don't seem to help much.
\\

\begin{center}
\textbf{Perturbed - 60\% Similarity}    
\end{center}

I have been having headaches for a while now. They are usually on the left side of my head and are very painful. I also get nausea, vomiting, and sensitivity to light and sound. I have tried taking over-the-counter \textbf{\textit{agony}} relievers, but they don't seem to help much.
\\

\begin{center}
\textbf{Perturbed - 50\% Similarity}    
\end{center}

I have been having headaches for a while now. They are usually on the left side of my head and are very painful. I also get nausea, vomiting, and sensitivity to light and sound. I have tried taking over-the-counter \textbf{\textit{agony}} relievers, but they don't seem to \textbf{\textit{assistance}} much.
\\

\begin{center}
\textbf{Perturbed - 40\% Similarity}    
\end{center}

I have been having headaches for a while now. They are usually on the left side of my head and are very painful. I also get nausea, vomiting, and sensitivity to light and sound. I have \textbf{\textit{strived}} taking over-the-counter \textbf{\textit{agony}} relievers, but they don't \textbf{\textit{appears}} to \textbf{\textit{aiding}} much.
\\

\begin{center}
\textbf{Perturbed - 30\% Similarity}    
\end{center}

I have been \textbf{\textit{assuming}} headaches for a while now. They are usually on the left \textbf{\textit{sides}} of my head and are very painful. I also get nausea, vomiting, and sensitivity to light and sound. I \textbf{\textit{ai}} \textbf{\textit{strived}} taking over-the-counter \textbf{\textit{agony}} relievers, but they don't \textbf{\textit{appears}} to \textbf{\textit{aiding}} much.
\vspace{5pt}
\hrule
\end{table*}

\clearpage
\onecolumn
\section{Bounding Penalized Spearman's Footrule Maximum Distance}\label{apn:spearman}
Recall that Spearman's footrule possesses a maximum total distance (when the list is completely inverted) of $\lfloor \frac{|\mathbf{A}|^2}{2}\rfloor$ for $\mathbf{|A|} = \mathbf{|B|}, \; \mathbf{A} \cap \mathbf{B} = \mathbf{A}$ with an individual element having at most $|\mathbf{A}| - 1$ possible distance.

\noindent Given a penalty, $p \geq 1$,  for disjoint elements (i.e. there exists $e \in \mathbf{A}\; s.t \;e \notin \mathbf{B}$), we may have a total possible distance that exceeds the nonadjusted footrule.
\newline

\noindent Let $\mathbf{A}, \mathbf{B}$ be lists that are possibly disjoint and without loss of generality assume $\mathbf{|A|} \geq \mathbf{|B|} $. 
\newline

\noindent As the largest possible individual distance is $|\mathbf{A}| - 1$, assume $p \geq \mathbf{|A|} - 1$. Then trivially, the maximum distance is $p \mathbf{|A|}$.
\newline

\noindent Now assume $p < \mathbf{|A|} - 1$. The individual distances for a completely inverted list follow the patterns of: $\mathbf{|A|} - 1, \mathbf{|A|} - 3, \mathbf{|A|} - 5 ... 1, 1, ... \mathbf{|A|} - 5, \mathbf{|A|} - 3, \mathbf{|A|} - 1$ for $\mathbf{|A|}$ even, and $\mathbf{|A|} - 1, \mathbf{|A|} - 3, \mathbf{|A|} - 5 ... 0 ... \mathbf{|A|} - 5, \mathbf{|A|} - 3, \mathbf{|A|} - 1$ for $\mathbf{|A|}$ odd.
\newline

\noindent We can view the new maximum penalty as the sum of two components: $p_g$, the individual distances greater than or equal to the penalty, and $p_l$, those less than the penalty.
\newline

\noindent As $p < \mathbf{|A|} - 1$, there exists some distance $d \; s.t \; p \leq d \leq \mathbf{|A|} - 1$. Let this $d$ be located at the index $i$ within the pattern $\mathbf{|A|} - 1, \mathbf{|A|} - 3, \;...$ , where $\mathbf{|A|} - 1$ is index $1$. The size of this subset of the above pattern is $\mathbf{|A|} - 2i$ and so its sum can be given as $\lfloor \frac{({|\mathbf{A}| - 2i})^2}{2}\rfloor$.
\newline

\noindent Then $p_g = \lfloor \frac{|\mathbf{A}|^2}{2}\rfloor - \lfloor \frac{({\mathbf{{|A|}} - 2i )}^2}{2}\rfloor$. And since $p \geq d$, then $p_l = p (\mathbf{{|A|}}-2i) -  \lfloor \frac{({|\mathbf{A}| - 2i})^2}{2}\rfloor$.
\newline 

\noindent So, the maximum distance for $p < \mathbf{|A|} - 1$  is given by $p_g + p_l =  
 \lfloor \frac{|\mathbf{A}|^2}{2}\rfloor + p (\mathbf{{|A|}} - 2i) -  2\lfloor \frac{({|\mathbf{A}| - 2i})^2}{2}\rfloor$.
\newline

\end{document}